%% file: 00_main.tex
\documentclass[letterpaper, 10pt, journal, twoside]{IEEEtran}
\IEEEoverridecommandlockouts

\usepackage{amsmath}
\usepackage{mathtools}
\usepackage{cite}
\usepackage{amsmath,amssymb,amsfonts}
\usepackage[font=small, labelfont=bf]{caption}
\usepackage[ruled, lined, linesnumbered, commentsnumbered, longend, noend]{algorithm2e}

\usepackage{dblfloatfix}
\usepackage{hyperref}
\usepackage{algorithmic}
\usepackage{graphicx}
\usepackage{textcomp}
\usepackage{breqn}
\usepackage{gensymb}
\usepackage{cuted}
\usepackage{capt-of}
\usepackage{booktabs}
\usepackage{diagbox}
\usepackage{makecell}
\usepackage{multirow}
\usepackage{multicol}

\usepackage{pifont}
\usepackage[table]{xcolor}
\newcommand{\cmark}{\ding{51}}
\newcommand{\xmark}{\textcolor{gray}{\ding{55}}}
\newcommand{\omark}{\textcolor{gray}{\ding{108}}}

\usepackage{fancyhdr} 
\def\BibTeX{{\rm B\kern-.05em{\sc i\kern-.025em b}\kern-.08em
    T\kern-.1667em\lower.7ex\hbox{E}\kern-.125emX}}

\definecolor{pink}{RGB}{255, 192, 203}

\definecolor{darkgreen}{RGB}{83, 199, 34}

\title{\LARGE \bf \textit{DoublyAware:} Dual Planning and Policy Awareness for Temporal Difference Learning in Humanoid Locomotion}

\author{Khang Nguyen$^{1}$, An T. Le$^{2}$, Jan Peters$^{2,3,4}$, and Minh Nhat Vu$^{5,6}$ 
\thanks{$^{1}$University of Texas at Arlington, Texas, USA}
\thanks{$^{2}$Intelligent Autonomous Systems Lab, TU Darmstadt, Germany} 
\thanks{$^{3}$German Research Center for AI (DFKI), SAIROL, Darmstadt, Germany} 
\thanks{$^{4}$Hessian.AI, Darmstadt, Germany}
\thanks{$^{5}$Automation \& Control Institute (ACIN), TU Wien, Vienna, Austria} 
\thanks{$^{6}$Austrian Institute of Technology (AIT) GmbH, Vienna, Austria}
\thanks{E-mails: \href{mailto:khang.nguyen8@mavs.uta.edu}{\text{khang.nguyen8@mavs.uta.edu}}, \href{mailto:minh.vu@ait.ac.at}{\text{minh.vu@ait.ac.at}}.}
}

\begin{document}

\maketitle
\thispagestyle{empty}
\pagestyle{empty}

\begin{abstract}
    Achieving robust robot learning for humanoid locomotion is a fundamental challenge in model-based reinforcement learning (MBRL), where environmental stochasticity and randomness can hinder efficient exploration and learning stability. The environmental, so-called aleatoric, uncertainty can be amplified in high-dimensional action spaces with complex contact dynamics, and further entangled with epistemic uncertainty in the models during learning phases. In this work, we propose \textit{DoublyAware}, an uncertainty-aware extension of Temporal Difference Model Predictive Control (TD-MPC) that explicitly decomposes uncertainty into two disjoint interpretable components, \textit{i.e.}, planning and policy uncertainties. To handle the planning uncertainty, \textit{DoublyAware} employs conformal prediction to filter candidate trajectories using quantile-calibrated risk bounds, ensuring statistical consistency and robustness against stochastic dynamics. Meanwhile, policy rollouts are leveraged as structured informative priors to support the learning phase with Group-Relative Policy Constraint (GRPC) optimizers that impose a group-based adaptive trust-region in the latent action space. This principled combination enables the robot agent to prioritize high-confidence, high-reward behavior while maintaining effective, targeted exploration under uncertainty. Evaluated on the \texttt{HumanoidBench} locomotion suite with the Unitree 26-DoF H1-2 humanoid, \textit{DoublyAware} demonstrates improved sample efficiency, accelerated convergence, and enhanced motion feasibility compared to RL baselines. Our simulation results emphasize the significance of structured uncertainty modeling for data-efficient and reliable decision-making in TD-MPC-based humanoid locomotion learning.
\end{abstract}

\input{01_introduction}
\input{02_related_work}
\input{03_methodology}
\input{04_evaluation}
\input{05_conclusions}

\bibliographystyle{IEEEtran}
\bibliography{IEEEabrv, 09_references}

\end{document}

%% file: 01_introduction.tex
\vspace{-4pt}
\section{Introduction}

In model-based reinforcement learning (MBRL) for humanoid locomotion learning, uncertainty is a central concern for ensuring robustness and safe behaviors, particularly for high-dimensional, complex, whole-body coordination, where observations and dynamics can be noisy and stochastic \cite{radosavovic2023learning, radosavovic2024real}. As humanoid robots need to explore and interact with world dynamics, they must adaptively reason about two fundamentally distinct sources of uncertainty: one inherent to the environment and one arising from limitations in learned policy knowledge. Therefore, in this work, we identify two complementary forms of uncertainty to tackle this problem:
\begin{itemize}
    \item Planning uncertainty arises from the stochasticity in the local sampling-based trajectory optimizers, such as Model Predictive Control \cite{kouvaritakis2016model}, in the planning phase.
    \item Policy uncertainty stems from the learned policy network's incomplete knowledge due to the unexplored action-state space during the learning phase.
\end{itemize}

\begin{figure}[t]
    \vspace{2pt}
    \centering    \includegraphics[width=1.0\linewidth]{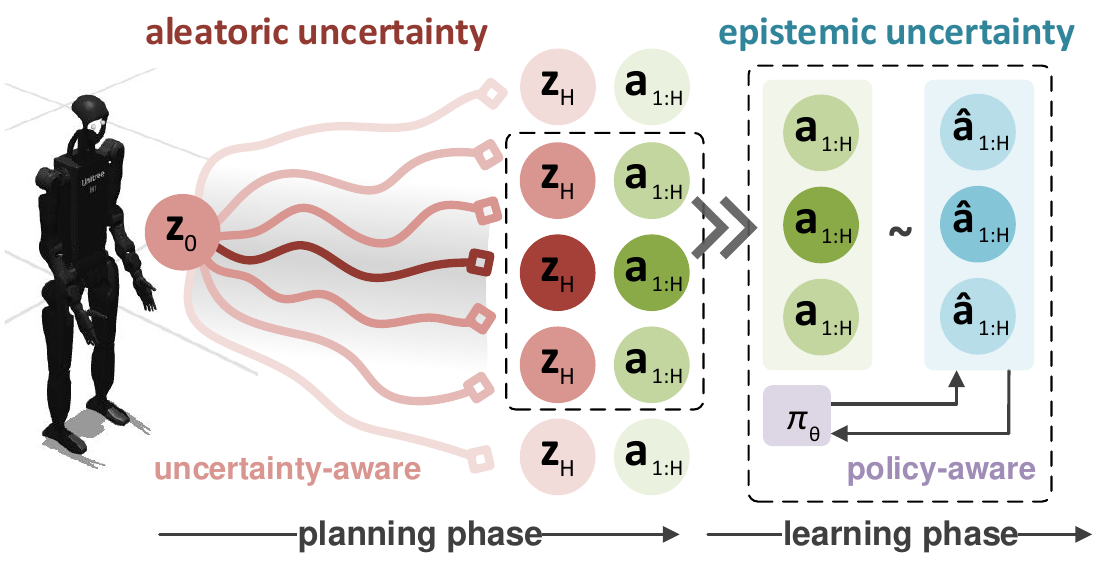}
    \vspace{-15pt}
    \caption{\textbf{Overview of \textit{DoublyAware}:} Disjoint uncertainty decomposition in TD-MPC frameworks and refinements for each component in planning and learning phases for robust humanoid locomotion.}
    \vspace{-20pt}
    \label{fig:doubleaware_concept}
\end{figure}

Planning uncertainty maps to \textit{aleatoric uncertainty}, which is induced by the environment randomness together with the system dynamics (\textit{e.g.}, ground contact, observation noises, and multi-modal nature of feasible movement trajectories). Such uncertainty cannot be eliminated even with augmented data, as it naturally reflects the stochasticity of the world itself. Meanwhile, policy uncertainty is akin to \textit{epistemic uncertainty}, where the model is being trained yet to know about the world due to its internal training experience, which can be reduced through further exploration, learning, and sophisticated policy-aware optimization. 

The concept of aleatoric and epistemic uncertainty can be dated back to prior works in machine learning \cite{shaker2020aleatoric, hullermeier2021aleatoric}. Vast adaptations have been made for further investigation in learning the world dynamics and control drifts under the influence of uncertainty in classification \cite{shaker2020aleatoric} for feature selection and autonomous driving \cite{distelzweig2024stochasticity, hagedorn2025learning} for trajectory prediction under the influence of control uncertainty and world dynamics. In the scope of learning for control, Temporal Difference Model Predictive Control (TD-MPC) \cite{hansen2022temporal, hansen2023td} has shown its excellence at short-horizon decision-making through real-time trajectory optimization with TD-learning, enabling more flexible and scalable behavior learning for MBRL-based techniques. Still, these methods frequently suffer from poor sample efficiency and unstable policy updates \cite{chua2018deep, argenson2020model} and notably remain vulnerable to planning compound errors and learning biases issues that are particularly pronounced in high-dimensional control settings \cite{hansen2022temporal, hansen2023td, lin2025td}.

To address this dual challenge, we proposed \textit{DoublyAware}, a planning- and policy-aware TD-MPC-based method for humanoid locomotion learning. Specifically, our approach explicitly decomposes aleatoric planning and epistemic learning uncertainties and solves them distinctively, as illustrated in Fig. \ref{fig:doubleaware_concept}. Leveraging conformal prediction theory \cite{vovk2005algorithmic}, \textit{DoublyAware} integrates conformal quantile filtering to select suitable candidate trajectories robustly, and uses them as informative priors to modulate agents into a high-reward learning space as a planning-aware solution. Furthermore, inspired by recent advancements in large language models, \textit{DoublyAware} incorporates group-relative policy optimizers \cite{shao2024deepseekmath} in the learning phase with an adaptive trust-region as policy-aware learning. Our contributions are threefold:
\begin{enumerate}
    \item We outline the decomposition of overall uncertainty in MBRL into planning and policy uncertainty, and solve them distinctively instead of framing them as one.
    \item We integrate the planning-aware mechanism with the policy-aware learning optimizer, enabling uncertainty-calibrated trajectory filtering followed by policy rollouts used as informative learnable priors.
    \item We evaluate \textit{DoublyAware}'s performance on locomotion tasks in \texttt{HumanoidBench} \cite{sferrazza2024humanoidbench}, showcasing its improvements compared to baseline methods regarding learning speed and kinodynamically feasible motions.
\end{enumerate}

%% file: 02_related_work.tex
\vspace{-4pt}
\section{Related Work}

\textbf{Temporal-Difference Model Predictive Control:} Humanoid locomotion is one of the most complex control systems, stemming from its high-dimensional continuous action spaces, inherently unstable dynamics, and complex interactions with the environment \cite{peters2003reinforcement}. TD-MPC has shown promise in addressing these challenges by uniting the short-horizon optimization capabilities of MPC with the sample-efficient, value-driven learning of RL. Prior works on this \cite{sikchi2022learning, hansen2022temporal, hansen2023td} have demonstrated that incorporating TD learning into MPC frameworks enables flexible value function learning without relying on handcrafted cost functions. Building upon this direction, TD-MPC2 \cite{hansen2023td} extends the original TD-MPC \cite{hansen2022temporal} by introducing scalable latent world models tailored for continuous control to mitigate error accumulation and improve planning stability, where previously even minor errors can quickly lead to destabilized motion as a result. By merging TD learning with MPC-style planning, these frameworks enhance sample efficiency and offer adaptability in high-dimensional control. However, these methods fail when encountering more complex tasks as their uncertainty compounds over time for an extended task execution period. In this work, we further investigate and solve the planning and learning uncertainty distinctively that might lead to poor performance and non-feasible behavior of the TD-MPC framework for humanoid control, especially for locomotion tasks.

\textbf{Uncertainty-Aware Robot Planning:} Recent developments in uncertainty-aware robotics have emphasized the role of conformal prediction and information-theoretic decomposition to enhance planning robustness. In trajectory and motion planning, conformal prediction offers statistical guarantees through distribution-free calibration, making it a natural fit for high-risk, multimodal robotic tasks. Prior works have integrated conformal methods into learned manifold learning \cite{kuleshov2018conformal, kiyani2024conformal}, enabling model-agnostic risk assessment for learned representations. This research has been extended to motion planning under dynamic uncertainty, where adaptive conformal frameworks improve safety and feasibility \cite{doulaconformal, sun2023conformal, dixit2023adaptive, lindemann2023safe, yang2023safe, lekeufack2024conformal, sheng2024safe}. Other approaches have explored handling distribution shifts during policy learning via conformal mechanisms \cite{huang2024conformal}, while others target high-dimensional control for teleoperation through confidence-aware policy mappings \cite{zhao2024conformalized}. Complementing these, Stochasticity in Motion introduces an entropy-based decomposition of trajectory uncertainty into aleatoric and epistemic terms, formalizing their roles in motion prediction and emphasizing their implications for safe downstream planning. Unlike previous approaches, our work directly applies conformal prediction to latent trajectory selection between policy-guided priors and stochastic trajectories to alleviate exploration uncertainty during planning, offering a solution for effectively tackling the aleatoric part of the TD-MPC framework.

\textbf{Policy-Aware Optimization for Robot Learning:} Policy mismatching poses a core challenge in robot control, particularly for humanoid locomotion, where off-policy methods are highly susceptible to discrepancies between the actions rolled out by the learned policy and the targets generated by temporal-difference updates. The misalignment and bootstrapping error accumulation often lead to compounding inaccuracies and poor generalization performance \cite{kumar2019stabilizing, kumar2020conservative}. Offline RL approaches have significantly addressed this by learning from fixed datasets. Several methods mitigate distributional shift by explicitly regularizing the policy toward expert demonstrations \cite{kumar2019stabilizing, fujimoto2021minimalist}, while others leverage importance sampling to correct for distributional mismatch in value estimation \cite{fujimoto2019off, peng2019advantage}. Similarly, in-sample learning techniques \cite{garg2023extreme, kostrikov2021offline} avoid out-of-distribution actions by constraining updates to observed data, implicitly ensuring policy reliability. This challenge is equally framed in MBRL; for example, LOOP \cite{sikchi2022learning} introduces actor regularization to inject conservatism into the planning process and stabilize learning. Departing from prior work, our method enforces distributional consistency directly on the policy prior in latent space, without modifying the underlying planner, allowing for more flexible planning while maintaining stability and enabling fast policy adaptation. In brief, our work extends these principles by incorporating algorithmic stability and data efficiency through Group-Relative Policy Optimization (GRPO) \cite{shao2024deepseekmath} with an explicit trust-region constraint for policy optimization during the learning phase.

%% file: 03_methodology.tex
\vspace{-4pt}
\section{Planning- \& Policy-Aware Temporal Difference Learning}
\label{sec:methodology}

\begin{figure*}[t]
    \vspace{1pt}
    \centering
    \includegraphics[width=1.0\linewidth]{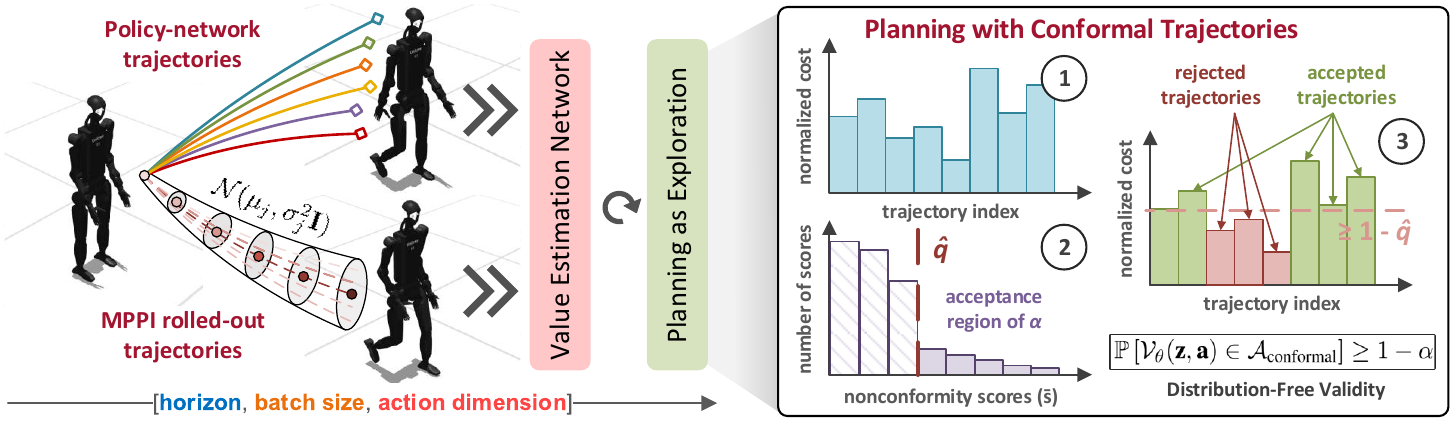}
    \caption{\textbf{Uncertainty-Aware Planning for Humanoid Locomotion:} At each planning step, two sets of trajectories are sampled from the policy network and MPPI planner. The policy network provides prior-guided trajectories that reflect learned behavior, while MPPI samples explore around a sampling distribution $\mathcal{N}(\mu_j, \sigma_j^2 \mathbf{I})$ with $j$ as the number of iterations. Each iteration contains a batch of trajectories for all action dimensions along the predictive planning horizon. (1) These candidate trajectories are then evaluated using the TD-based cost function, (2) assigned normalized nonconformity scores $ \bar{s} $, (3) and filtered via a conformal quantile threshold $\hat{q}$ to retain only trajectories within the empirical $(1 - \alpha)$ prediction set. These conformal latent trajectories ensure statistically reliable planning under policy model error and guarantee quality explorative behavior for sequential decision-making while learning humanoid locomotion tasks.}
    \vspace{-17pt}
    \label{fig:planclt_framework}
\end{figure*}

\subsection{Vanilla Planning with Model-Predictive Control}
Humanoid locomotion tasks can be formulated as infinite-horizon Markov Decision Processes (MDPs), defined by the tuple $\mathcal{M} = (\mathcal{S}, \mathcal{A}, \rho, r, \gamma)$, where $\mathcal{S}$ denotes the state space, $\mathcal{A}$ is the action space, $\rho: \mathcal{S} \times \mathcal{A} \rightarrow \mathcal{S}$ represents the transition function, $r: \mathcal{S} \times \mathcal{A} \rightarrow \mathbb{R}$ is the reward function, and $\gamma \in (0, 1]$ is the discount factor. The objective $J^{\pi}$ is to learn the parameters $\theta$ for the policy network $\Pi_{\theta}: \mathcal{S} \rightarrow \mathcal{A}$ that enables the robot to continuously generate optimal actions, maximizing the expected discounted cumulative reward over a trajectory, $\xi$, as a sequence of states and actions:
\begin{equation}
    J^{\pi} = \mathbb{E}_{\xi \sim \Pi_{\theta}} \left[ \sum_{t=0}^{\infty} \gamma^{t} r(\textbf{s}_{t}, \textbf{a}_{t}) \right], \text{ } \xi = \begin{bmatrix}
        \textbf{s}_{0} & \textbf{s}_{1} & \dots \\
        \textbf{a}_{0} & \textbf{a}_{1} & \dots \\
    \end{bmatrix}
    \label{eq:objective}
\end{equation}
with each action $\mathbf{a}_t$ is sampled from the policy $\Pi_{\theta}(\mathbf{s}_{t})$, and each subsequent state $\mathbf{s}_t$ is determined by $\rho(\mathbf{s}_{t-1}, \mathbf{a}_{t-1})$ based on the previous state $\mathbf{s}_{t-1}$ and action $\mathbf{a}_{t-1}$.

Hansen \textit{et al.} \cite{hansen2022temporal, hansen2023td} introduced an approach that integrates MPPI \cite{williams2016aggressive} as a local trajectory optimizer for short-horizon planning within a learned latent dynamics model. During the planning phase, the action sequences of length $H$ are sampled as latent trajectories from the learned dynamics model, and the cumulative return $\phi_{\xi}$ of each sampled trajectory $\xi$ is calculated using a trained value estimator $\mathcal{V}_{\theta}(\textbf{z}_{t}, \textbf{a}_{t})$, as the composition of $R_{\theta}(\textbf{z}_{t}, \textbf{a}_{t})$ and $Q_{\theta} (\textbf{z}_{H}, \textbf{a}_{H})$ as follows:
\begin{equation}
    \mathcal{V}_{\theta}(\textbf{z}_{t}, \textbf{a}_{t}) \coloneqq \phi_{\xi} = \sum_{t=0}^{H-1} \gamma^{t} R_{\theta}(\textbf{z}_{t}, \textbf{a}_{t}) + \gamma^{H} Q_{\theta} (\textbf{z}_{H}, \textbf{a}_{H}),
    \label{eq:traj_update}
\end{equation}
where $\textbf{z}_{t} = h_{\theta}(\textbf{s}_{t})$ is the latent representation that selectively captures the relevant dynamics of the state $\textbf{s}_{t}$, rather than all observation dimensions, $\textbf{z}_{t} = d_{\theta}(\textbf{s}_{t-1}, \textbf{a}_{t-1})$ represents the next latent state under the latent dynamics $d_{\theta}$, $\hat{r}_{t} = R_{\theta}(\textbf{s}_{t}, \textbf{a}_{t})$ and $\hat{q}_{t} = Q_{\theta}(\textbf{s}_{t}, \textbf{a}_{t})$ denote reward and value estimators, and $ \textbf{a}_{t} \sim \mathcal{N}(\mu_{t}, \sigma_{t}^2 \mathbf{I})$ describes the MPPI sampling distributions:
\begin{equation}
    \mu_{j} = \eta \sum_{i=1}^k \Omega_i \xi_i^*, \quad \sigma_{j}^2 = \eta \sum_{i=1}^k \Omega_i(\xi_i^* - \mu^j)^2,
    \label{eq:traj_moments}
\end{equation}
where $\Omega_i = \exp(\tau \phi_{\xi, i}^*)$, $\tau$ is a temperature parameter, $\eta$ represents the normalizing term with respect to the coefficients $\Omega_i$, and $\xi_i^*$ denotes the $i^{\text{th}}$ of the latent trajectory corresponding to return estimate $\phi_{\xi}^*$.  In the MBRL-based robot learning, Eq. \ref{eq:traj_update} can therefore be called as an $H$-step look-ahead policy, which iteratively maximizes the first step's costs.

\vspace{-4pt}
\subsection{Planning as Exploration with Conformal Trajectories}
To improve planning exploration probabilistically safely, we integrate conformal prediction into the MPPI planner, as depicted in Alg. \ref{alg:plan_clt}. Specifically, conformal prediction allows the underlying planner to choose ``\textit{statistically-significant}'' trajectories as candidates without assuming any parametric form of the return value/cost distribution.

The planning procedure in TD-MPC2 \cite{hansen2023td} incorporates two distinct sources of candidate latent trajectories (Fig. \ref{fig:planclt_framework}):
\begin{itemize}
    \item The first set is sampled from the current policy network by simulating trajectories under the learned world model, which acts as prior knowledge by proposing trajectories that reflect the agent’s learned behavior so far. These trajectories serve as a warm start for planning, reducing dependence on purely random sampling and encouraging consistency during training.
    \item The second set is generated via the MPPI planner, which aims to explore the action space heuristically and re-weights them based on expected cumulative rewards, thus refining the action distribution toward higher-value trajectories through the learning process. 
\end{itemize}

Using policy-guided and MPPI samples improves exploration and stability by balancing prior-driven guidance with adaptive search. Nevertheless, both sources might fall into sub-optimality, where policy rollouts may propagate error-prone behavior during training, while MPPI may overfit to inaccuracies in the model dynamics. Thus, an uncertainty-aware selection mechanism for the MPPI planner is needed to calibrate and filter trajectories based on empirical plausibility.

We compute the conformal scores to quantify the agreement between candidate trajectories and the prior trajectories from the policy network. Denote $\mathcal{A}_\pi = \{\textbf{a}^{\pi, (i)}_{1:H}\}_{i=1}^{N_\pi}$ as the $N_{\pi}$ prior trajectories sampled from the policy with $v_{1:H}^{\pi, (i)}$ are their evaluated costs, $\mathcal{A}_{\text{mppi}} = \{\textbf{a}^{(i)}_{1:H}\}_{i=1}^N$ as the $N$ candidate trajectories generated via MPPI with $v_{1:H}^{(i)}$ as their corresponding costs. With Eq. \ref{eq:traj_update}, the nonconformity scores can be directly computed from the latent trajectories' costs:
\begin{equation}
    \bar{s}^{(i)} = 1 - \eta \mathcal{V}_{\theta}\left(\textbf{z}^{(i)}_{1:H}, \textbf{a}^{(i)}_{1:H}\right) \text{ } \forall \textbf{a} \in \mathcal{A} = \{ \mathcal{A}_{\pi} \cup \mathcal{A}_{\text{mppi}} \}
    \label{eq:nonconformity_score}
\end{equation}
where the scores are computed as the normalized cost functions across all actions $\textbf{a}^{(i)}_{1:H}$ along the planning horizon $H$. Note that normalized costs are computed to align with the original concept of softmaxes, in the range of 0 to 1, in classical conformal classification problems.

As both sets are exchangeable and the TD-based cost function $\mathcal{V}_{\theta}(\textbf{z}, \textbf{a})$ is used to estimate the conformal scores, the conformal prediction set $\mathcal{A}_{\text{conformal}}$ satisfies the marginal coverage guarantee, so-called distribution-free validity:
\begin{equation}
    \textstyle
    1 - \alpha + \frac{1}{(N_{\pi} + N) + 1} \geq \mathbb{P}\left[\mathcal{V}_{\theta}(\textbf{z}, \textbf{a}) \in \mathcal{A}_{\text{conformal}}\right] \geq 1 - \alpha.
    \label{eq:conformal_validity}
\end{equation}

\setlength{\textfloatsep}{4pt}
\begin{algorithm}[t]
    \caption{Planning with Conformal Trajectories}
    \label{alg:plan_clt}
    \begin{normalsize}
        \DontPrintSemicolon
        \SetKwInOut{KwIn}{Input}
        \SetKwFunction{FMainTrain}{plan}
        \SetKwProg{Pn}{function}{}{}
        \KwIn{$H$: planning horizon, $J$: number of iterations, $\mathcal{N}(\mu_{0}, \sigma_{0}^{2}\mathbf{I})$: initial distribution for MPPI, \\$\textbf{z}_{t}$: latent represent. at time $t$, $\alpha$: error rate, \\$N_{\pi}$: number of prior trajectories by $\pi_{\theta}$, \\$N$: number of MPPI-sampled trajectories} 
        \KwOut{$\textbf{a}_t$: action sampled from $\mathcal{N}(\mu_J, \sigma_J^2 \mathbf{I})$}
        \Pn{\FMainTrain{$H$, $J$, $\textbf{z}_{t}$, $\alpha$, $\mathcal{N}(\mu_{0}, \sigma_{0}^{2}\mathbf{I})$, $N$, $N_\pi$}}{
            \While{planning}{
                \For{$j = 1, \dots, J$}{
                    $\mathcal{A}_{\pi} \gets \{ \mathbf{a}_{i}^{\pi} \}_{i=1}^{N_{\pi}} \sim \pi_{\theta}$ \\
                    $\mathcal{A}_{\text{mppi}} \gets \{ \mathbf{a}_{i}^{\text{mppi}} \}_{i=1}^{N} \sim \mathcal{N}(\mu_{j-1}, \sigma_{j-1}^2 \mathbf{I})$ \\
                    \For{$\mathbf{a}_i \in  \mathcal{A} = \{ \mathcal{A}_{\pi} \cup \mathcal{A}_{\text{mppi}} \}$}{
                        $v_i \gets 0$, $\textbf{z}_0 \gets \textbf{z}_t$ \\
                        \For{$t = 0, \dots, H-1$}{
                            $v_i \gets v_i + \gamma^t R_{\theta}(\textbf{z}_t, \textbf{a}_{i,t})$ \\
                            $\textbf{z}_{t+1} \gets d_{\theta}(\textbf{z}_t, \textbf{a}_{i,t})$ \\
                        }
                        $v_i \gets v_i + \gamma^H Q_{\theta}(\textbf{z}_H, \textbf{a}_{i,H})$ {\small (Eq. \ref{eq:traj_update})} \\
                    }
                    \textcolor{gray}{\small // compute nonconformity scores from costs} \\
                    $\bar{s}_i \gets 1 - \texttt{normalize}(v_i)$  {\small (Eq. \ref{eq:nonconformity_score})} \\
                    $\hat{q} \gets \texttt{quantile}\left(\{\bar{s}_i\}, 1 - \alpha\right)$ {\small (Eq. \ref{eq:conformal_quantile})} \\
                    $\mathcal{A}_{\text{conformal}} \gets \{ i \in \mathcal{A} \mid \bar{s}_i \leq \hat{q} \}$  {\small (Eq. \ref{eq:conformal_set})} \\
                    \textcolor{gray}{\small // update parameters for next planning step} \\
                    $\mu_j, \sigma_j \gets \texttt{update}(\{\mathbf{a}_i\}_{i \in \mathcal{A}_{\text{conformal}}})$  {\small (Eq. \ref{eq:traj_moments})} \\
                }
            }
            \Return $a_{\tau} \sim N(\mu, \sigma^2 \mathbf{I}) \text{ } \textbf{for }\tau = t, \dots, t+H$
        }
    \end{normalsize}
\end{algorithm}

With the nonconformity scores $\bar{s}^{(i)}$ from Eq. \ref{eq:nonconformity_score}, we construct the prediction set of candidate trajectories that conform to the statistical properties of the calibration set. The key idea is to select a quantile threshold $\hat{q}$ such that a fixed proportion $(1 - \alpha)$ of calibration trajectories achieve scores less than or equal to this threshold. Formally, we define it:
\begin{equation}
    \hat{q} = \texttt{quantile} \left[ \left\{ \bar{s}^{(i)} \right\}_{i=1}^{N_\pi + N}, (1 - \alpha) \right],
    \label{eq:conformal_quantile}
\end{equation}
where $\alpha \in [0,1)$ is the pre-defined risk level that controls the desired coverage. Intuitively, $\hat{q}$ defines a level set of conformity: any candidate MPPI latent trajectory with a score $\bar{s}^{(j)} \le \hat{q}$ is deemed statistically compatible with the teacher prior $\mathcal{A}_\pi$. This leads to the formal definition of the conformal prediction set that is in Eq. \ref{eq:conformal_validity}, described as:
\begin{equation}
    \mathcal{A}_{\text{conformal}} = \left\{ \textbf{a}^{(j)}_{1:H} \in \mathcal{A} \mid \bar{s}^{(j)} \le \hat{q} \right\},
    \label{eq:conformal_set}
\end{equation}
which serves as a filtered subset of high-confidence trajectories. The conformal set $\mathcal{A}_{\text{conformal}}$ admits finite-sample validity guarantees under the assumption that the calibration scores $\{ \bar{s}^{(i)} \}$ and the candidate scores $\{ \bar{s}^{(j)} \}$ are exchangeable for any $i$ and $j$ within the union set size. Thus, $\mathcal{A}_{\text{conformal}}$ contains the best trajectories with probability at least $1-\alpha$, which satisfies Eq. \ref{eq:conformal_validity}. Moreover, in online learning, the value estimator is improved over time as it learns periodically. 

\begin{figure*}[t]
    \vspace{2pt}
    \centering
    \includegraphics[width=1.0\linewidth]{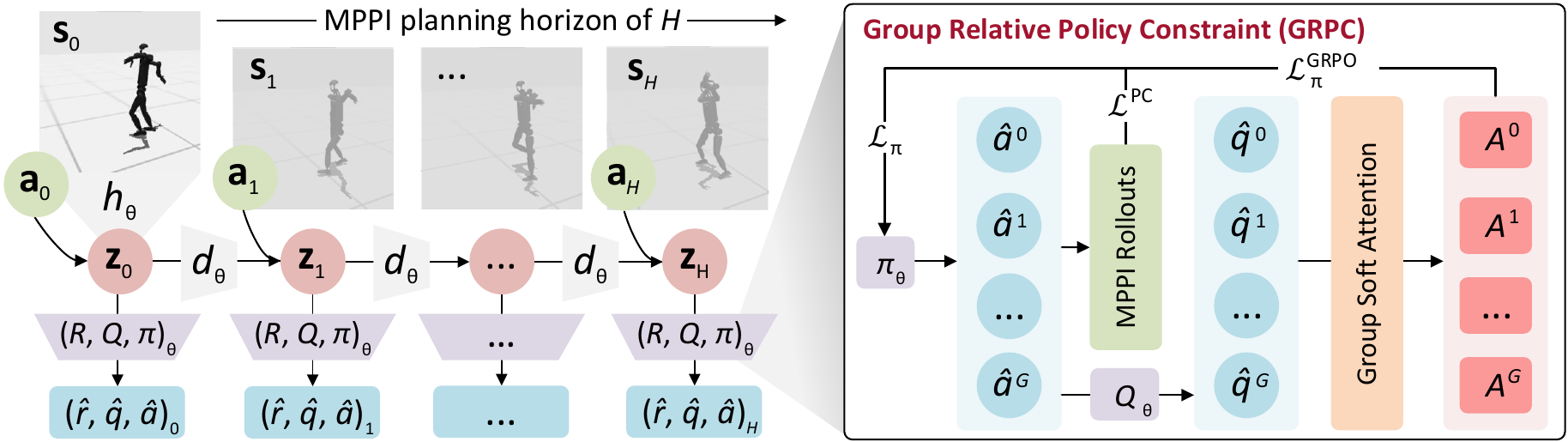}
    \caption{\textbf{Policy-Aware Learning for Humanoid Locomotion:} Given an initial state $\textbf{s}_{0}$, it is encoded into a latent representation $\textbf{z}_{0}$ using the encoder $h_{\theta}$. A latent dynamics model $d_{\theta}$ then iteratively predicts future latent states $\textbf{z}_{t+1}$ over a planning horizon of $H$ steps, conditioned on actions $\textbf{a}_{t}$ and current latent states $\textbf{z}_{t}$. Throughout this horizon, reward estimates, $Q$-values, and policy outputs are generated via learned networks $R_{\theta}$, $Q_{\theta}$, and $\pi_{\theta}$, respectively, supporting trajectory optimization in latent space guided by temporal-difference objectives. For each state, groups of sampled actions are rolled out and evaluated through $Q$-values, which are transformed into softmax-weighted advantage scores $A^{g}$ per action group $g$. These scores inform the GRPC objective, promoting the selection of high-value actions while reducing policy variance. A trust-region is enforced through a KL divergence penalty between the current policy and its MPPI-based prior, thereby regularizing residual policy learning and bounding divergence from prior behavior to ensure stable updates.}
    \vspace{-12pt}
    \label{fig:tdgrpc_framework}
\end{figure*}

For this reason, we suppose that the TD-based cost function $\mathcal{V}_{\theta}(\textbf{z}, \textbf{a})$ is an imperfectly taught value estimator (\textit{i.e.}, a weak teacher in conformal prediction theory \cite{vovk2005algorithmic}) that reveals informative conformity scores only along a teaching schedule $\mathcal{L} = \{n_k\}_{k \ge 1} \subset \mathbb{N}$. If the teaching schedule satisfies the sub-linearity condition $\lim_{k \to \infty} ({n_k}/{n_{k-1})} = 1$, the conformal quantile $\hat{q}$ in Eq. \ref{eq:conformal_quantile}, constructed from calibration scores $\{\bar{s}^{(i)}\}$, admits asymptotic weak validity. That is, the prediction set $\mathcal{A}_{\text{conformal}}$ defined in Eq. \ref{eq:conformal_set} satisfies:
\begin{equation}
    \liminf_{\mathcal{L} \to \infty} \mathbb{P} \left[ \bar{s}^{\text{test}} \leq \hat{q} \right] \ge 1 - \alpha.
    \label{eq:asymptotic_validity_weak_priors}
\end{equation}

This means that this asymptotic validity under weak priors (Eq. \ref{eq:asymptotic_validity_weak_priors}) holds even if the cost signal from $\mathcal{V}_{\theta}(\textbf{z}, \textbf{a})$ is imperfect or inconsistent, as long as the prior model improves.

Indeed, Eq. \ref{eq:traj_update} is a black-box oracle, and no distributional assumptions are made beyond exchangeability within this context. Therefore, this conformal filtering preserves the model-agnostic nature while seamlessly integrating with latent dynamics and value-based planning, highlighting that the delay in learning the value estimator is allowed.

\subsection{Learning with Group Relative Policy Constraint}
We adopt and improve GRPO \cite{shao2024deepseekmath} to enhance action group-based explicit advantage estimation, enhancing entropy-regularized policy gradient methods by leveraging group-wise action comparisons, enabling the policy to learn from relative action preferences rather than relying on those priors. In standard actor-critic methods, policy gradients are scaled by absolute values or advantage estimates, which may be sensitive to estimation errors. These limitations become especially pronounced in long-horizon tasks; GRPO addresses this issue by constructing a relative preference distribution across sampled groups. Mathematically, at each state $\textbf{s}_{i}$, a set of $G$ actions $\{\textbf{a}_{i}^{1}, \dots, \textbf{a}_{i}^{G}\}$ is sampled, and their $Q$-values $\{q_{i}^{1}, \dots, q_{i}^{G}\}$ are computed, which are used to compute the group-based soft attention advantage scores:
\begin{equation}
    A_{i}(\mathbf{q}) = \frac{\exp(q_{i} / \tau)}{\sum_{g=1}^G \exp(q_{i}^{g} / \tau)},
    \label{eq:weights}
\end{equation}
where $\mathbf{q} = Q_\theta(\textbf{s}, \textbf{a}_i)$ denotes the estimates, and $\tau$ is a temperature parameter and $0 \leq A_{i}(\cdot) \leq 1$ as its property.

As $\{\textbf{a}_i^g\}_{g=1}^G$ is a group of $G$ actions sampled from a policy $\pi_\theta(\textbf{s})$ at the state $\textbf{s}$ with $\hat{r}_i = r_{\theta}(\textbf{s}, \textbf{a}_i)$ and $\hat{q}_i = Q_{\theta}(\textbf{s}, \textbf{a}_i)$ are the reward and estimated values, respectively, we assume that $||\nabla_\theta \log \pi_\theta(\textbf{a} \mid \textbf{s})|| = C$, and $\hat{q}_i$, $\hat{r}_i$ are bounded above with $\forall \textbf{a} \in \mathcal{A}$ and $\forall \textbf{s} \in \mathcal{S}$. Therefore, we obtain:
\begin{equation}
    \mathrm{Var} \left[ \nabla_{\theta} \mathcal{L}_{\text{softmax}} \right] \leq \mathrm{Var} \left[ \nabla_{\theta} \mathcal{L}_{\text{std-norm}} \right],
    \label{eq:variance_inequality}
\end{equation}
with $\mathcal{L}_{\text{softmax}}$ and $\mathcal{L}_{\text{std-norm}}$ are the softmax-based and standard normalized advantage losses, respectively. The variance of gradient of $\mathcal{L}_{\text{softmax}}$ is thus smaller than that of $\mathcal{L}_{\text{std-norm}}$:
\begin{equation}
    ||\nabla_{\theta} \mathcal{L}_{\text{softmax}}|| \text{ \small is bounded, }  ||\nabla_{\theta} \mathcal{L}_{\text{std-norm}}||\text{ \small is unbounded} 
    \label{eq:bounded_loss}
\end{equation}
yields more stable policy updates at some constant $C$ that asymptotically bounds $||\nabla_{\theta} \log \pi_\theta(\textbf{a} \mid \textbf{s})||$. Two keys favor softmax-based over normalized advantages. First, their outputs lie between 0 and 1, limiting the impact of outliers. Meanwhile, normalized advantages induce large magnitudes under noise, leading to high-variance gradients. Second, policy gradients scale with the advantage values. The gradient steps will be disproportionately unstable if the advantage is large or small. Therefore, softmax-based advantages smooth out extreme values and act like a soft attention mechanism, giving more stable policy updates during training episodes.

With the group relative weights in Eq. \ref{eq:weights} and based on Eq. \ref{eq:variance_inequality} and Eq. \ref{eq:bounded_loss}, the improved GRPO objective is defined as:
\begin{equation}
    \mathcal{L}_{\pi}^{\text{GRPO}}(\theta) = \frac{1}{G} \sum_{i=1}^{G} A_{i}(\textbf{q}) \log \pi_{\theta}(\textbf{a}_i \mid \textbf{s})
    \label{eq:grpo}
\end{equation}
where $\mu_k$ denotes the behavior policy at $k^{th}$ iteration from $\mathcal{D}$ in Eq. \ref{eq:traj_moments}. The KL constraint ensures the updated policy remains within a trust region of $\pi$. The overall policy objective combines the trust-region matching with Eq. \ref{eq:grpo}:
\begin{equation}
    \mathcal{L}_{\pi}(\theta) = \frac{1}{G} \sum_{i=1}^{G} A_{i}(\textbf{q}) \log \pi_{\theta}(\textbf{a}_i \mid \textbf{s})) + \beta \log \mu(\textbf{a} \mid \textbf{s}),
    \label{eq:policy_objective}
\end{equation}
where $\beta$ is a weighting coefficient controlling the penalty strength, the second term of Eq. \ref{eq:policy_objective} imposes a residual-style regularization as a trust-region for policy optimization \cite{schulman2015trust}.

Meanwhile, the latent dynamics $d_{\theta}$, encoder $h_{\theta}$, reward network $R_{\theta}$, and value network $Q_{\theta}$ are concurrently optimized by the following model objective:
\begin{subequations}
    \begin{align}
        \mathcal{L}&(\theta;\text{ }\xi_{i}) = 
        \left\| d_\theta(\mathbf{z}_i, \mathbf{a}_i) - h_\theta(\mathbf{s}_{i+1}) \right\|_2^2 \label{eq:latent_consistency} \\
        &+ \left\| R_\theta(\mathbf{z}_i, \mathbf{a}_i) - r_i \right\|_2^2 \label{eq:reward_consistency} \\
        &+ \left\| Q_\theta(\mathbf{z}_i, \mathbf{a}_i) - \left[r_i + \gamma Q_\theta(\mathbf{z}_{i+1}, \pi_\theta(\mathbf{z}_{i+1}))\right] \right\|_2^2 \label{eq:value_consistency}
    \end{align}
    \label{eq:model_objective}
    \vspace{-12pt}
\end{subequations}

\setlength{\textfloatsep}{4pt}
\begin{algorithm}[t]
    \caption{Learning with Group-Relative Constraint}
    \label{alg:td_grpc}
    \begin{normalsize}
        \DontPrintSemicolon
        \SetKwInOut{KwIn}{Input}
        \SetKwInOut{KwOut}{Output}
        \SetKwFunction{FMainTrain}{learn}
        \SetKwProg{Pn}{function}{}{}
        \KwIn{$T:$ trajectory length, $H$: planning horizon, \\$G$: number of groups, $\mathcal{D}$: latent buffer, \\$S$: number of iterations} 
        % \KwOut{}
        \Pn{\FMainTrain{$T$, $H$, $G$, $\mathcal{D}$, $S$}}{
            \While{learning}{
                \For{$t = 0, \dots, T$}{
                    $\textbf{a}_t \sim \Pi_{\theta}(h_{\theta}(\textbf{s}_t))$ \\
                    $(\textbf{s}_{t + 1}, r_t) \sim \mathcal{P}(\textbf{s}_t, \textbf{a}_t), \mathcal{R}_{\theta}(\textbf{s}_t, \textbf{a}_t)$ \\
                    $\mathcal{D} \leftarrow \mathcal{D} \cup (\textbf{s}_t, \textbf{a}_t, r_t, \textbf{s}_{t + 1})$ \\
                }
                \For{step = 0, \dots, S}{
                    $\{\textbf{s}_t, \textbf{a}_t, r_t, \textbf{s}_{t + 1}\}_{t : t+H}^{g} \sim \mathcal{D}$ {\small \textbf{for}} \textit{G} {\small groups} \\
                    $\mu^G, \sigma^G = \texttt{\small compute\_moments}(\textbf{a}_t)$ {\small (Eq. \ref{eq:traj_moments})} \\
                    $\textbf{z}_t \gets h_{\theta}(\textbf{s}_t)$ \textbf{if} $\textbf{s}_t$ {\small is the first observation} \\
                    \For{$i = t, \dots, t + H$}{
                        $\textbf{z}_{i + 1} \gets d_{\theta}(\textbf{z}_i, \textbf{a}_i)$ {\small (Eq. \ref{eq:latent_consistency})} \\
                        $\hat{r}_i \gets R_{\theta}(\textbf{z}_i, \textbf{a}_i)$ {\small (Eq. \ref{eq:reward_consistency})}\\
                        \textcolor{gray}{\small // group sampling \& policy constraint} \\
                        \For{$g = 1, \dots, G$ }{ 
                            $\widehat{\textbf{a}}_i^g \sim \pi_{\theta}(\textbf{z}_i)$ \\
                            $\varepsilon = (\widehat{\textbf{a}}_i^g - \mu^{G}) / \sigma^{G}$ \\
                            $\widehat{\textbf{a}}_i^g \leftarrow \texttt{\small{threshold}}(\widehat{\textbf{a}}_i^g, \varepsilon)$ \\
                            $\hat{q}_i^g = Q_{\theta}(\textbf{z}_i, \widehat{\textbf{a}}_i^g)$ {\small (Eq. \ref{eq:value_consistency})} \\
                        }
                        $A_i^g = \texttt{\small{softmax}}(\widehat{\textbf{q}}^g)$ {\small (Eq.~\ref{eq:weights})} \\
                        $\mathcal{L}_{\pi}^{(i)} = \frac{1}{G} \sum_{g=1}^{G} A_i^g \log \pi_\theta(\widehat{\textbf{a}}_i^g \mid \textbf{z}_i)$
                    }
                    $\mathcal{L}_\pi = \frac{1}{H} \sum_{i=t}^{t+H} \left[ \mathcal{L}_{\pi}^{(i)} + \beta \mathcal{L}_{\mathrm{KL}} \right]$ {\small (Eq. \ref{eq:policy_objective})} \\
                    $\theta \leftarrow \theta - \eta \nabla_{\theta} \mathcal{L}_{\pi}$
                }
            }
        }
    \end{normalsize}
\end{algorithm}

The training procedure with TD learning with GRPC at each short horizon for long-horizon locomotion tasks is summarized in Alg. \ref{alg:td_grpc} and is described visually in Fig. \ref{fig:tdgrpc_framework}.

%% file: 04_evaluation.tex
\section{Simulation Results \& Ablation Studies}
\label{sec:evaluations}

\begin{figure*}[t]
    \vspace{2pt}
    \centering
    \includegraphics[width=0.99\linewidth]{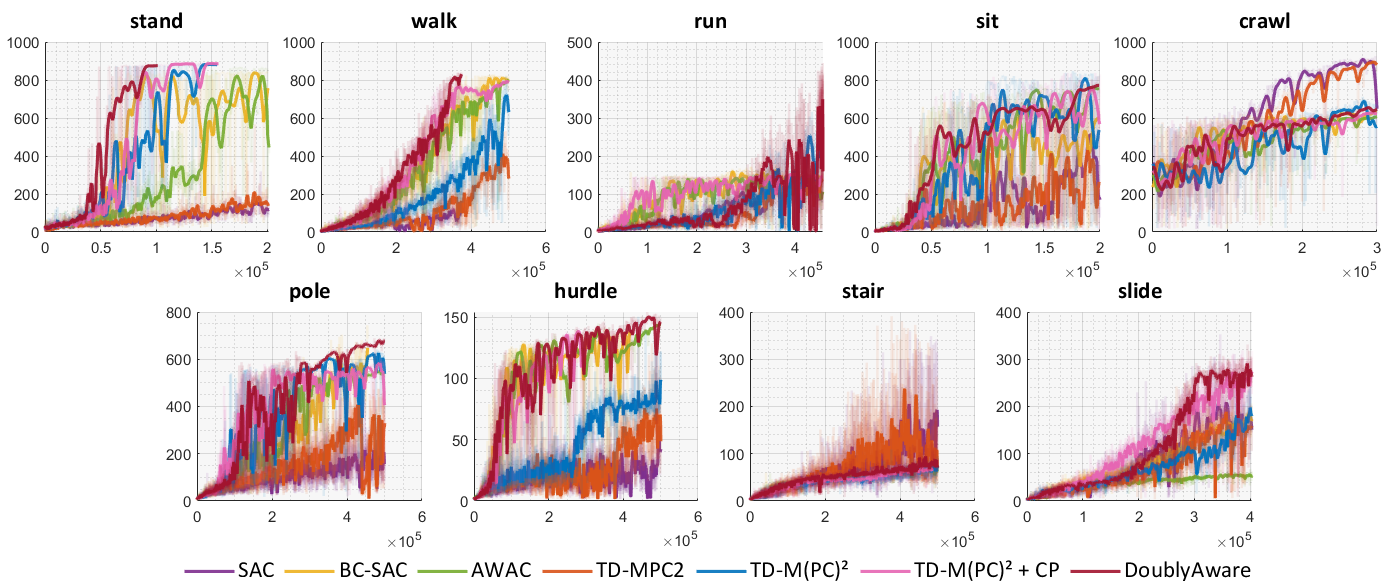}
    \vspace{-2pt}
    \caption{\textbf{Episode Returns of \textit{DoublyAware} and Baselines on H1–2 in Locomotion Tasks:} \textit{DoublyAware} achieves rapid convergence over others in standing, walking, sitting, navigating through poles, hurdling, and sliding tasks, while it performs worse in more complex tasks such as crawling and stair-climbing, which require high-dimensional whole-body coordination. In general, \textit{DoublyAware} shows slightly better data-efficiency than TD-M(PC)$^2$, BC-SAC, and AWAC and significantly better sampling-efficiency than SAC and TD-MPC2. 
    }
    \vspace{-16pt}
    \label{fig:returns_comparison}
\end{figure*}

To assess our proposed method's performance, we train \textit{DoublyAware} on \textbf{the Unitree 26-DoF H1-2 humanoid}, comprised of two legs of 12 DoFs and two arms of 14 DoFs. The torso joint is locked to eliminate unnecessary body-turning actions. The hands are included in the robot's model during training to account for their mass, ensuring that the learned policies take both hands into consideration, even though the robot is not involved in manipulation tasks. We evaluate \textit{DoublyAware}'s performance against SAC \cite{haarnoja2018soft}, BC-SAC \cite{lu2023imitation}, AWAC \cite{nair2020awac}, TD-MPC2 \cite{hansen2023td}, TD-M(PC)$^2$ \cite{lin2025td} on the locomotion tasks in \texttt{HumanoidBench} \cite{sferrazza2024humanoidbench}. Nine tasks include standing, walking, running, sitting on a chair, crawling through a tunnel, navigating through standing poles, hurdling, stair-climbing, and walking over slides. Our evaluations are to answer the following questions:
\begin{enumerate}
    \item Does \textit{DoublyAware} achieve superior reward convergence compared to existing methods?
    \item Can \textit{DoublyAware} successfully solve tasks that demand long-horizon, whole-body coordination?
    \item Whether conformal trajectory planning and group-relative policy constraint learning mutually benefit?
    \item How does the motion generated by \textit{DoublyAware} compare qualitatively to those from other baselines?
\end{enumerate}

Across the experiments, the evaluated algorithms are set with the default starting pose of the H1-2. The hyperparameters include the planning horizon of $3$, batch size of $256$, action dimension of $26$, learning rate of $0.0003$, and number of prior trajectories of $24$ on an AMD Ryzen 9 7950X3D CPU and an NVIDIA RTX 4090 GPU. Additionally, we use a group number of $3$ for group-based policy optimization and an error rate of $0.05$ for conformal trajectory planning.

\vspace{-7pt}
\subsection{Episode Returns of Locomotion Tasks}
We report the episode return comparisons across all evaluated methods on the \texttt{HumanoidBench} locomotion tasks in Fig. \ref{fig:returns_comparison}. On foundational tasks such as standing, walking, running, and sitting, \textit{DoublyAware} demonstrates significantly faster convergence than competing approaches. Specifically, the H1-2 humanoid achieves upright standing in fewer than $100,000$ training iterations, walking in approximately $300,000$ iterations, running in $450,000$ iterations, and sitting in $200,000$ iterations. In contrast, baseline methods such as SAC, BC-SAC, AWAC, TD-MPC2, and TD-M(PC)$^2$ show their difficulties achieving comparable performance on walking and running within the same training budget.

On more complex tasks requiring precise whole-body coordination, such as navigating around standing poles, hurdling over obstacles, and walking across inclined slides, \textit{DoublyAware} maintains a consistent performance advantage. Notably, it surpasses a reward threshold of 600 on the pole navigation task within $500,000$ iterations, achieves a reward of $150$ on hurdling, and reaches around $300$ on walking over slides. While other methods eventually learn these tasks, they exhibit slower convergence rates and higher variability in performance. On the most challenging tasks, such as crawling and stair-climbing, \textit{DoublyAware} underperforms relative to other methods in terms of reward acquisition within the same number of training iterations. Overall, \textit{DoublyAware} shows the training efficiency compared to the competing baselines.

\begin{figure*}[t]
    \vspace{2pt}
    \centering
    \includegraphics[width=1.0\linewidth]{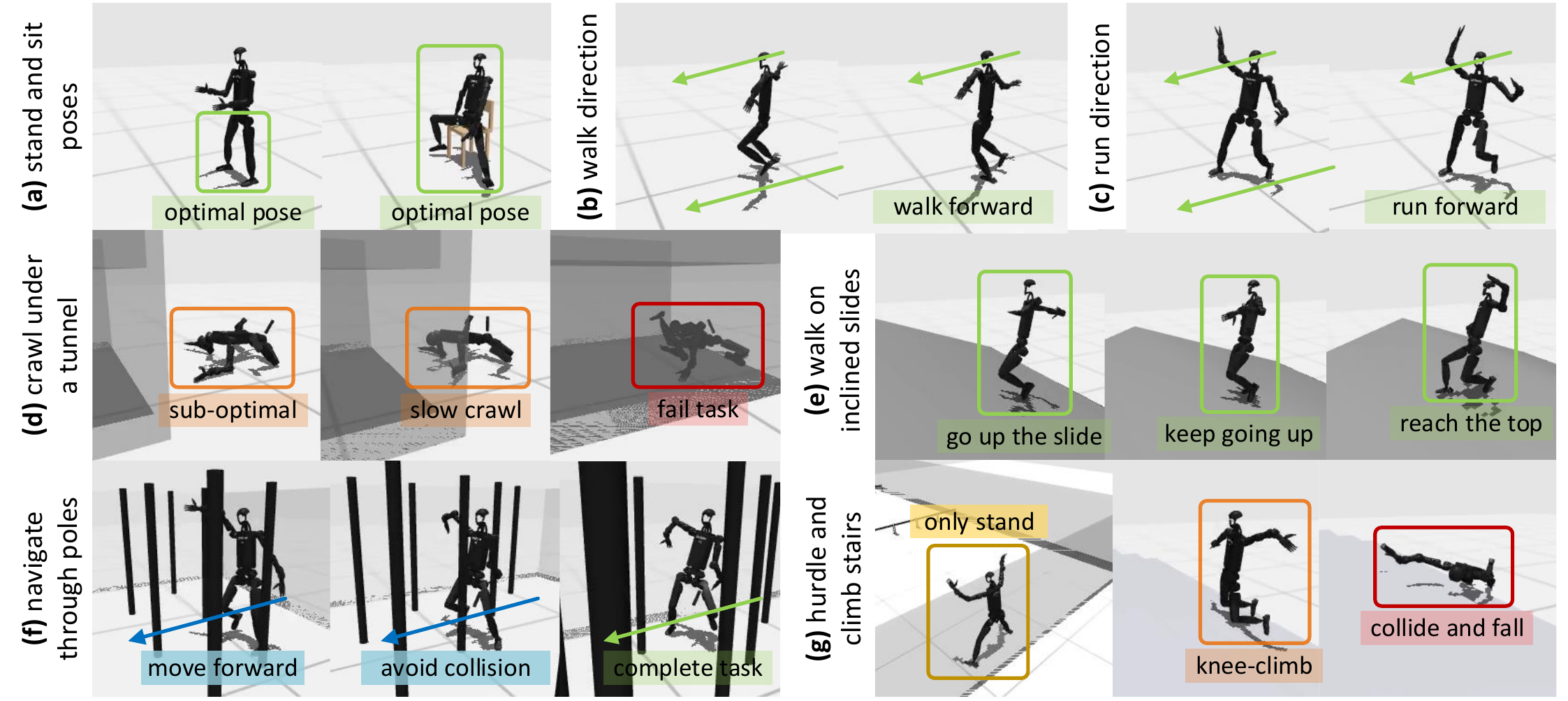}
    \caption{\textbf{Qualitative Results of H1-2 in Locomotion Tasks:} (a) \textit{Standing \& Sitting Poses:} \textit{DoublyAware} enables H1-2 to complete these tasks with appropriate leg and hand poses. (b) and (c) \textit{Walking \& Running Direction:} H1-2 can walk and run in the forward direction when trained with \textit{DoublyAware}, unlike when being trained with other algorithms: TD-M(PC)$^2$ induces walking/running backward, others generate dynamically-infeasible motions. (d) \textit{Crawling under A Tunnel:} \textit{DoublyAware} unable to teach the robot to crawl through the tunnel, where TD-M(PC)$^2$ and the ablation methods also fail. Only SAC and TD-MPC2 can accomplish this task. (e) \textit{Walking on Inclined Slides:} The H1-2 can reach the top of the slide-like hill faster than other baselines when learning with \textit{DoublyAware}. (f) \textit{Navigating through Poles:} Again, \textit{DoublyAware} generates feasible motions for the robot to move forward and avoid collision; other methods fail on this task. and (g) \textit{Hurdling \& Stair-Climbing:} \textit{DoublyAware} is unable to teach the H1-2 to accomplish these challenging tasks, so do other methods.}
    \vspace{-16pt}
    \label{fig:qualitative_results}
\end{figure*}

\begin{table*}[!b]
    \vspace{-5pt}
    \centering
    \caption{Solving ability of SAC \cite{haarnoja2018soft}, BC-SAC \cite{lu2023imitation}, AWAC \cite{nair2020awac}, TD-MPC2 \cite{hansen2023td}, TD-M(PC)$^2$ \cite{lin2025td}, and our method, \textit{DoublyAware}, of locomotion tasks on H1-2 in \texttt{HumanoidBench} \cite{sferrazza2024humanoidbench}:~\cmark~for tasks that are solved sufficiently,~\omark~for tasks that need additional mild refinements for success, and~\xmark~for tasks that need further intensive learning of whole-body and selective dynamics features.}
    \vspace{-2pt}
    \begin{tabular}{l ccccccccc}
        \toprule
        Method / Task & \texttt{stand} & \texttt{walk} & \texttt{run} & \makecell{\texttt{sit on} \\ \texttt{chair}} & \makecell{\texttt{crawl under} \\ \texttt{tunnel}} & \makecell{\texttt{navigate} \\ \texttt{through poles}} & \texttt{hurdle} &  \makecell{\texttt{climb} \\ \texttt{stairs}} & \makecell{\texttt{walk on} \\ \texttt{inclined slide}} \\
        \midrule \midrule
        SAC \cite{haarnoja2018soft} & \xmark & \xmark & \xmark & \xmark & \cmark & \xmark & \xmark & \xmark & \omark \\
        BC-SAC \cite{lu2023imitation} & \xmark & \xmark & \xmark & \omark & \omark & \omark & \xmark & \xmark & \xmark \\
        AWAC \cite{nair2020awac} & \omark & \xmark & \xmark & \omark & \omark & \omark & \xmark & \xmark & \xmark \\
        TD-MPC2 \cite{hansen2023td} & \xmark & \xmark & \xmark & \omark & \cmark & \xmark & \xmark & \xmark & \omark \\
        TD-M(PC)$^2$ \cite{lin2025td} & \cmark & \omark & \omark & \omark & \omark & \omark & \xmark & \xmark & \omark \\
        \midrule
        \textbf{\textit{DoublyAware}} & \cmark & \cmark & \cmark & \cmark & \omark & \cmark & \xmark & \xmark & \cmark \\
        \bottomrule
    \end{tabular}
    \label{tab:solving_ability}
    % \vspace{-16pt}
\end{table*}

\vspace{-4pt}
\subsection{Ablation Studies on Uncertainty-Aware Modules}
As also shown in Fig. \ref{fig:returns_comparison}, we conduct an ablation analysis to examine whether conformal trajectory prediction (CP) and group-relative policy constraint (GRPC) offer complementary benefits when integrated into the TD-MPC framework. We compare three variants: TD-M(PC)$^2$ (blue) -- which is plain baseline, TD-M(PC)$^2$ + CP (pink), and \textit{DoublyAware} (red) -- which combines both CP and GRPC.

Across most locomotion tasks, \textit{DoublyAware} consistently outperforms the ablated variants in final performance and sample efficiency. For example, in tasks such as standing, walking, navigating through poles, hurdling, and walking over slides, \textit{DoublyAware} reaches peak returns faster and more stably than the other two. Although it underperforms in complex coordination tasks like crawling and stair-climbing, TD-M(PC)$^2$ + CP performs better in these settings, benefiting from CP’s uncertainty-aware planning -- yet it lacks generalization across broader task domains. In general, these results suggest that CP and GRPC are mutually beneficial, each addressing complementary aspects of the problem: CP improves robustness and uncertainty-awareness in trajectory exploration, and GRPC enhances policy-aware policy optimization for learning. Their integration in \textit{DoublyAware} leads to a more data-efficient and robust locomotion policy.

\vspace{-4pt}
\subsection{Visualization of Sequential Behaviors}
Beyond task completion, we study the qualitative results of generated behavior across learning algorithms. Fig. \ref{fig:qualitative_results} shows the performance of H1-2 in a range of locomotion tasks. 

In Fig. \ref{fig:qualitative_results}a, \textit{DoublyAware} enables the robot to achieve stable and plausible poses for standing and sitting, demonstrating coordinated control of the legs and limbs. In contrast, baseline methods exhibit unstable or implausible configurations. Fig. \ref{fig:qualitative_results}b and Fig. \ref{fig:qualitative_results}c both show that \textit{DoublyAware} consistently results in coherent walking and running in the forward direction, while TD-M(PC)$^2$ and other methods induce backward locomotion. For Fig. \ref{fig:qualitative_results}e, in the slide-walking task, \textit{DoublyAware} guides the robot to ascend the incline progressively and reach the top.

In the more challenging scenarios, such as Fig. \ref{fig:qualitative_results}d, crawling through a tunnel, \textit{DoublyAware} fails to complete the task, showing sub-optimal postures and stalled progress, similar to baseline methods, which also fail to solve this task. Only SAC and TD-MPC2 can complete this task sufficiently. Fig. \ref{fig:qualitative_results}f shows that \textit{DoublyAware} improves spatial awareness in navigating between standing poles, generating trajectories that avoid collisions while maintaining forward progress, unlike competing methods that get stuck or misstep. Lastly, in Fig. \ref{fig:qualitative_results}g, DoublyAware and also other baselines fail to teach the H1-2 hurdle and climb stairs. In general, these results highlight that while \textit{DoublyAware} significantly improves performance on many tasks compared to its competitors.

We summarize solving abilities across all locomotion tasks for all algorithms in Table \ref{tab:solving_ability}. Based on both quantitative and qualitative results analyzed, the summary shows an empirical improvement when training the H1-2 with \textit{DoublyAware}. In specific, \textit{DoublyAware} successfully solves most locomotion tasks, including standing, walking, running, sitting on a chair, navigating through standing poles, and walking on an inclined slide. However, \textit{DoublyAware} and others cannot solve crawl, hurdle, and climb stairs tasks, which require more advanced whole-body coordination and dynamic skill refinement. Overall, these results highlight the robustness of \textit{DoublyAware} in diverse locomotion settings, with room for improvement in tasks demanding complex full-body movements. For more comprehensive results, the demonstration video can be seen at: \href{https://www.acin.tuwien.ac.at/f3c8/}{\small \texttt{https://www.acin.tuwien.ac.at/f3c8/}}. 

%% file: 05_conclusions.tex
\section{Conclusions}

This work presents \textit{DoublyAware}, an uncertainty-aware extension of TD-MPC tailored for robust and sample-efficient humanoid locomotion. By decomposing uncertainty into disjoint planning and policy components, our method enables principled reasoning and mitigation in planning and learning phases. Conformal quantile filtering ensures statistically grounded trajectory selection under aleatoric uncertainty, while GRPC with adaptive trust-region regularization promotes stable and policy-aware learning under epistemic uncertainty. Our evaluations on \texttt{HumanoidBench} demonstrate that \textit{DoublyAware} surpasses prior methods in convergence speed and motion quality across various whole-body locomotion tasks, highlighting the benefits of structured uncertainty modeling in complex, high-dimensional humanoid control settings.

%% file: 00_main.bbl
\begin{thebibliography}{10}
\providecommand{\url}[1]{#1}
\csname url@samestyle\endcsname
\providecommand{\newblock}{\relax}
\providecommand{\bibinfo}[2]{#2}
\providecommand{\BIBentrySTDinterwordspacing}{\spaceskip=0pt\relax}
\providecommand{\BIBentryALTinterwordstretchfactor}{4}
\providecommand{\BIBentryALTinterwordspacing}{\spaceskip=\fontdimen2\font plus
\BIBentryALTinterwordstretchfactor\fontdimen3\font minus \fontdimen4\font\relax}
\providecommand{\BIBforeignlanguage}[2]{{%
\expandafter\ifx\csname l@#1\endcsname\relax
\typeout{** WARNING: IEEEtran.bst: No hyphenation pattern has been}%
\typeout{** loaded for the language `#1'. Using the pattern for}%
\typeout{** the default language instead.}%
\else
\language=\csname l@#1\endcsname
\fi
#2}}
\providecommand{\BIBdecl}{\relax}
\BIBdecl

\bibitem{radosavovic2023learning}
I.~Radosavovic, T.~Xiao, B.~Zhang, T.~Darrell, J.~Malik, and K.~Sreenath, ``Learning humanoid locomotion with transformers,'' \emph{CoRR}, 2023.

\bibitem{radosavovic2024real}
------, ``Real-world humanoid locomotion with reinforcement learning,'' \emph{Science Robotics}, vol.~9, no.~89, p. eadi9579, 2024.

\bibitem{kouvaritakis2016model}
B.~Kouvaritakis and M.~Cannon, ``Model predictive control,'' \emph{Switzerland: Springer International Publishing}, vol.~38, no. 13-56, p.~7, 2016.

\bibitem{shaker2020aleatoric}
M.~H. Shaker and E.~H{\"u}llermeier, ``Aleatoric and epistemic uncertainty with random forests,'' in \emph{International Symposium on Intelligent Data Analysis}.\hskip 1em plus 0.5em minus 0.4em\relax Springer, 2020, pp. 444--456.

\bibitem{hullermeier2021aleatoric}
E.~H{\"u}llermeier and W.~Waegeman, ``Aleatoric and epistemic uncertainty in machine learning: An introduction to concepts and methods,'' \emph{Machine learning}, vol. 110, no.~3, pp. 457--506, 2021.

\bibitem{distelzweig2024stochasticity}
A.~Distelzweig, A.~Look, E.~Kosman, F.~Janjo{\v{s}}, J.~Wagner, and A.~Valada, ``Stochasticity in motion: An information-theoretic approach to trajectory prediction,'' \emph{arXiv preprint arXiv:2410.01628}, 2024.

\bibitem{hagedorn2025learning}
S.~Hagedorn, A.~Distelzweig, M.~Hallgarten, and A.~P. Condurache, ``Learning through retrospection: Improving trajectory prediction for automated driving with error feedback,'' \emph{arXiv preprint arXiv:2504.13785}, 2025.

\bibitem{hansen2022temporal}
N.~A. Hansen, H.~Su, and X.~Wang, ``Temporal difference learning for model predictive control,'' in \emph{International Conference on Machine Learning}.\hskip 1em plus 0.5em minus 0.4em\relax PMLR, 2022, pp. 8387--8406.

\bibitem{hansen2023td}
N.~Hansen, H.~Su, and X.~Wang, ``Td-mpc2: Scalable, robust world models for continuous control,'' in \emph{The Twelfth International Conference on Learning Representations}.

\bibitem{chua2018deep}
K.~Chua, R.~Calandra, R.~McAllister, and S.~Levine, ``Deep reinforcement learning in a handful of trials using probabilistic dynamics models,'' \emph{NeurIPS}, 2018.

\bibitem{argenson2020model}
A.~Argenson and G.~Dulac-Arnold, ``Model-based offline planning,'' \emph{arXiv preprint arXiv:2008.05556}, 2020.

\bibitem{lin2025td}
H.~Lin, P.~Wang, J.~Schneider, and G.~Shi, ``Improving td-mpc through policy constraint,'' \emph{arXiv preprint arXiv:2502.03550}, 2025.

\bibitem{vovk2005algorithmic}
V.~Vovk, A.~Gammerman, and G.~Shafer, \emph{Algorithmic learning in a random world}.\hskip 1em plus 0.5em minus 0.4em\relax Springer, 2005, vol.~29.

\bibitem{shao2024deepseekmath}
Z.~Shao, P.~Wang, Q.~Zhu, R.~Xu, J.~Song, X.~Bi, H.~Zhang, M.~Zhang, Y.~Li, Y.~Wu \emph{et~al.}, ``Pushing the limits of mathematical reasoning in open language models,'' \emph{arXiv preprint arXiv:2402.03300}, 2024.

\bibitem{sferrazza2024humanoidbench}
C.~Sferrazza, D.-M. Huang, X.~Lin, Y.~Lee, and P.~Abbeel, ``Humanoidbench: Simulated humanoid benchmark for whole-body locomotion and manipulation,'' \emph{arXiv preprint arXiv:2403.10506}, 2024.

\bibitem{peters2003reinforcement}
J.~Peters, S.~Vijayakumar, and S.~Schaal, ``Reinforcement learning for humanoid robotics,'' in \emph{Proceedings of the third IEEE-RAS international conference on humanoid robots}, 2003, pp. 1--20.

\bibitem{sikchi2022learning}
H.~Sikchi, W.~Zhou, and D.~Held, ``Learning off-policy with online planning,'' in \emph{CoRL}, 2022.

\bibitem{kuleshov2018conformal}
A.~Kuleshov, A.~Bernstein, and E.~Burnaev, ``Conformal prediction in manifold learning,'' in \emph{Conformal and Probabilistic Prediction and Applications}.\hskip 1em plus 0.5em minus 0.4em\relax PMLR, 2018, pp. 234--253.

\bibitem{kiyani2024conformal}
S.~Kiyani, G.~Pappas, and H.~Hassani, ``Conformal prediction with learned features,'' \emph{arXiv preprint arXiv:2404.17487}, 2024.

\bibitem{doulaconformal}
A.~Doula, T.~G{\"u}delh{\"o}fer, M.~M{\"u}hlh{\"a}user, and A.~S. Guinea, ``Conformal prediction for semantically-aware autonomous perception in urban environments,'' in \emph{8th Annual Conference on Robot Learning}.

\bibitem{sun2023conformal}
J.~Sun, Y.~Jiang, J.~Qiu, P.~Nobel, M.~J. Kochenderfer, and M.~Schwager, ``Conformal prediction for uncertainty-aware planning with diffusion dynamics model,'' \emph{Advances in Neural Information Processing Systems}, vol.~36, pp. 80\,324--80\,337, 2023.

\bibitem{dixit2023adaptive}
A.~Dixit, L.~Lindemann, S.~X. Wei, M.~Cleaveland, G.~J. Pappas, and J.~W. Burdick, ``Adaptive conformal prediction for motion planning among dynamic agents,'' in \emph{Learning for Dynamics and Control Conference}.\hskip 1em plus 0.5em minus 0.4em\relax PMLR, 2023, pp. 300--314.

\bibitem{lindemann2023safe}
L.~Lindemann, M.~Cleaveland, G.~Shim, and G.~J. Pappas, ``Safe planning in dynamic environments using conformal prediction,'' \emph{IEEE Robotics and Automation Letters}, vol.~8, no.~8, pp. 5116--5123, 2023.

\bibitem{yang2023safe}
S.~Yang, G.~J. Pappas, R.~Mangharam, and L.~Lindemann, ``Safe perception-based control under stochastic sensor uncertainty using conformal prediction,'' in \emph{2023 62nd IEEE Conference on Decision and Control (CDC)}.\hskip 1em plus 0.5em minus 0.4em\relax IEEE, 2023, pp. 6072--6078.

\bibitem{lekeufack2024conformal}
J.~Lekeufack, A.~N. Angelopoulos, A.~Bajcsy, M.~I. Jordan, and J.~Malik, ``Conformal decision theory: Safe autonomous decisions from imperfect predictions,'' in \emph{2024 IEEE International Conference on Robotics and Automation (ICRA)}.\hskip 1em plus 0.5em minus 0.4em\relax IEEE, 2024, pp. 11\,668--11\,675.

\bibitem{sheng2024safe}
S.~Sheng, P.~Yu, D.~Parker, M.~Kwiatkowska, and L.~Feng, ``Safe pomdp online planning among dynamic agents via adaptive conformal prediction,'' \emph{IEEE Robotics and Automation Letters}, 2024.

\bibitem{huang2024conformal}
H.~Huang, S.~Sharma, A.~Loquercio, A.~Angelopoulos, K.~Goldberg, and J.~Malik, ``Conformal policy learning for sensorimotor control under distribution shifts,'' in \emph{2024 IEEE International Conference on Robotics and Automation (ICRA)}.\hskip 1em plus 0.5em minus 0.4em\relax IEEE, 2024, pp. 16\,285--16\,291.

\bibitem{zhao2024conformalized}
M.~Zhao, R.~Simmons, H.~Admoni, and A.~Bajcsy, ``Conformalized teleoperation: Confidently mapping human inputs to high-dimensional robot actions,'' \emph{arXiv preprint arXiv:2406.07767}, 2024.

\bibitem{kumar2019stabilizing}
A.~Kumar, J.~Fu, M.~Soh, G.~Tucker, and S.~Levine, ``Stabilizing off-policy q-learning via bootstrapping error reduction,'' \emph{NeurIPS}, 2019.

\bibitem{kumar2020conservative}
A.~Kumar, A.~Zhou, G.~Tucker, and S.~Levine, ``Conservative q-learning for offline reinforcement learning,'' \emph{NeurIPS}, 2020.

\bibitem{fujimoto2021minimalist}
S.~Fujimoto and S.~S. Gu, ``A minimalist approach to offline reinforcement learning,'' \emph{NeurIPS}, vol.~34, pp. 20\,132--20\,145, 2021.

\bibitem{fujimoto2019off}
S.~Fujimoto, D.~Meger, and D.~Precup, ``Off-policy deep reinforcement learning without exploration,'' in \emph{ICML}, 2019.

\bibitem{peng2019advantage}
X.~B. Peng, A.~Kumar, G.~Zhang, and S.~Levine, ``Advantage-weighted regression: Simple and scalable off-policy reinforcement learning,'' \emph{arXiv preprint arXiv:1910.00177}, 2019.

\bibitem{garg2023extreme}
D.~Garg, J.~Hejna, M.~Geist, and S.~Ermon, ``Extreme q-learning: Maxent rl without entropy,'' \emph{arXiv preprint arXiv:2301.02328}, 2023.

\bibitem{kostrikov2021offline}
I.~Kostrikov, A.~Nair, and S.~Levine, ``Offline reinforcement learning with implicit q-learning,'' \emph{arXiv preprint arXiv:2110.06169}, 2021.

\bibitem{williams2016aggressive}
G.~Williams, P.~Drews, B.~Goldfain, J.~M. Rehg, and E.~A. Theodorou, ``Aggressive driving with model predictive path integral control,'' in \emph{2016 IEEE international conference on robotics and automation (ICRA)}.\hskip 1em plus 0.5em minus 0.4em\relax IEEE, 2016, pp. 1433--1440.

\bibitem{schulman2015trust}
J.~Schulman, S.~Levine, P.~Abbeel, M.~Jordan, and P.~Moritz, ``Trust region policy optimization,'' in \emph{International conference on machine learning}.\hskip 1em plus 0.5em minus 0.4em\relax PMLR, 2015, pp. 1889--1897.

\bibitem{haarnoja2018soft}
T.~Haarnoja, A.~Zhou, P.~Abbeel, and S.~Levine, ``Soft actor-critic: Off-policy maximum entropy deep reinforcement learning with a stochastic actor,'' in \emph{ICML}, 2018.

\bibitem{lu2023imitation}
Y.~Lu, J.~Fu, G.~Tucker, X.~Pan, E.~Bronstein, R.~Roelofs, B.~Sapp, B.~White, A.~Faust, S.~Whiteson \emph{et~al.}, ``Imitation is not enough: Robustifying imitation with reinforcement learning for challenging driving scenarios,'' in \emph{2023 IEEE/RSJ International Conference on Intelligent Robots and Systems (IROS)}.\hskip 1em plus 0.5em minus 0.4em\relax IEEE, 2023, pp. 7553--7560.

\bibitem{nair2020awac}
A.~Nair, A.~Gupta, M.~Dalal, and S.~Levine, ``Awac: Accelerating online reinforcement learning with offline datasets,'' \emph{arXiv preprint arXiv:2006.09359}, 2020.

\end{thebibliography}
